# *Efficient Commercial Bank Customer Credit Risk Assessment Based on LightGBM and Feature Engineering*

Sun Yanjie *
School of economics and management
University of Electronic Science and Technology of China
Chengdu, China
1789660136@qq.com

Gong Zhike*
School of economics and management
University of Electronic Science and Technology of China
Chengdu, China
2020150501004@std.uestc.edu.cn

Shi Quan
Information and Communication Engineering
University of Electronic Science and Technology of China
Chengdu, China
563832834@qq.com

Chen Lin†
School of economics and management
University of Electronic Science and Technology of China
Chengdu, China
chenlin2@uestc.edu.cn

***Abstract*—Effective control of credit risk is a key link in the steady operation of commercial banks. This paper is mainly based on the customer information dataset of a foreign commercial bank in Kaggle, and we use LightGBM algorithm to build a classifier to classify customers, to help the bank judge the possibility of customer credit default. This paper mainly deals with characteristic engineering, such as missing value processing, coding, imbalanced samples, etc., which greatly improves the machine learning effect. The main innovation of this paper is to construct new feature attributes on the basis of the original dataset so that the accuracy of the classifier reaches 0.734, and the AUC reaches 0.772, which is more than many classifiers based on the same dataset. The model can provide some reference for commercial banks' credit granting, and also provide some feature processing ideas for other similar studies.**

## I. INSTRUCTION

In the 1930s, J.M. Keynes proposed that "the amount of credit to any individual does not depend solely on the collateral and interest rate provided by the borrower, but also on the intention of the borrower and his position in the eyes of the bank,"[1] gradually forming the theory of credit rationing.

In recent years, the epidemic situation and grim international relations have led to increased downward pressure on China's economy and gradually expanded credit risks faced by commercial banks, which are the core of the financial system. By the end of the fourth quarter of 2021, the balance of non-performing loans of commercial banks was 2.8 trillion yuan, an increase of 13.5 billion yuan compared with the end of the previous quarter. In 2021, a total of 3.1 trillion yuan of non-performing assets in the banking sector were disposed of, exceeding 3 trillion yuan for two consecutive years[2]. Scientific assessment of banking credit risk plays an important role in the management of commercial banks.

Traditional commercial bank credit risk assessment mainly relies on the experience judgment of commercial bank practitioners, that is collected through offline customer personal information such as age, occupation, and historical records of default risk on person, but this kind of way often shortcomings such as low efficiency is difficult to guarantee, which is unable to meet the needs of the current banking credit business development. Therefore, the use of big data technology to achieve risk control is an area worthy of further research.

In previous research, common risk prediction methods have gradually developed, including artificial empirical judgment, mathematical analysis, and AI. Influenced by the development of information technologies such as machine learning and big data, this paper starts from the customer data of commercial banks and focuses on using machine learning algorithms to train models, such as LightGBM, to mine important risk factors affecting customer default and select appropriate parameters for model evaluation. Finally, the trained algorithm model is applied to the loan prediction of commercial banks to improve the credit risk evaluation mechanism and give more accurate suggestions for banks.

## II. DATA ANALYSIS

The customer sample dataset is derived from the Home Credit Default Risk project in Kaggle, with over 300,000 bank customer data. The dataset consists of 1 main table and multiple branch tables. The main table 'application_train' dataset includes 121 variables such as customer gender, car and real estate, number of defaults, etc. Each variable occupies one column of the table, while each row represents one credit application. X1, X2,..., X120 are the explanatory variables and Y is the target variable "TARGET". The main table will be used for the training of the model. Among the branch tables, 'bureau' is the credit history of each customer in other credit institutions; 'bureau_balance' is the record of each month in the credit report. 'previous_application' is the information of customers who have applied for Home Credit loans; 'POS_CASH_balance' is the monthly cash loans of customers; 'credit_card_balance' is the monthly credit card spending records of customers; 'installments_payments' is to get the customer's repayment history.

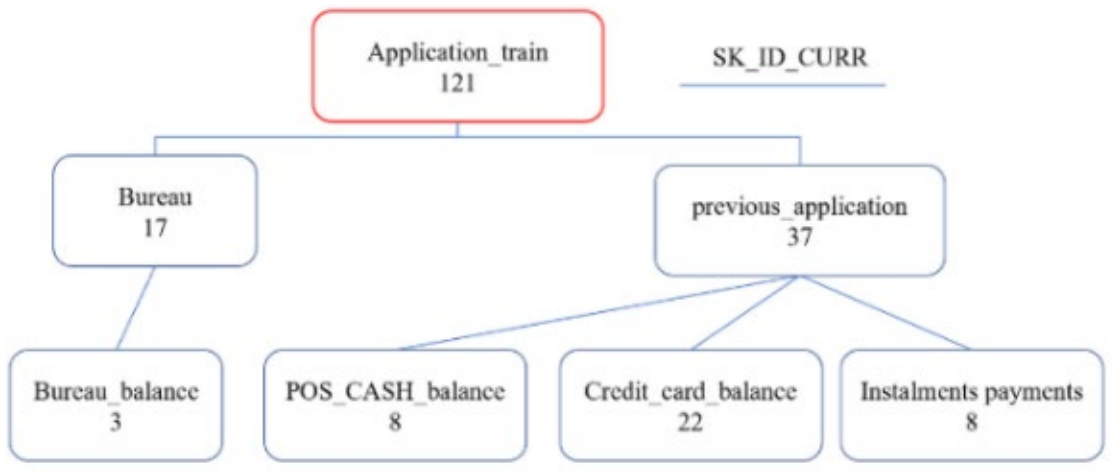

Fig. 1. Data set classification

*These authors contributed equally to this work. † Corresponding author

Some of variables are listed in Table Ⅰ.

Table I. EXPLANATION OF VARIABLES

| *Classification* | *Variable Name* | *Explain* | *Notes* |
|---|---|---|---|
| Basic Information | DAYS_BIRTH | Age at the time of user application | Numerical type |
| | OCCUPATION_TYPE | Occupation Type | Textual type |
| Assets | BASEMENTAREA_AVG | Size of residential apartments | Numerical type |
| | YEARS_BUILD_AVG | Age of the building | Numerical type |
| Credit Record | OBS_30_CNT_SOCIAL_CIRCLE | Number of times a customer is 30 days overdue | Numerical type |
| | AMT_CREDIT_SUM_DEBT | Current debt of credit institutions | Numerical type |
| Meaning Unclear | FLAG_DOCUMENT_2 | Does the user provide documentation | Numerical type |

We divided the features in the dataset into numeric and textual data according to their categories, with 104 numeric and 16 textual. The distribution of all feature variables was plotted to observe the distribution of the dataset.

Numerical data have both discrete and continuous data, and we found that there are two main distribution patterns by plotting the distribution histogram, i.e., normal distribution and unilateral distribution. There are 21 normal distributions, such as DAYS_BIRTH and HOUR_APPR_PROCESS_START, etc. There are 83 unilateral distributions, such as NONLIVINGAREA_MODE and AMT_INCOME_TOTAL, etc. Fig. 2. and Fig. 3. show some examples of the two distribution patterns.

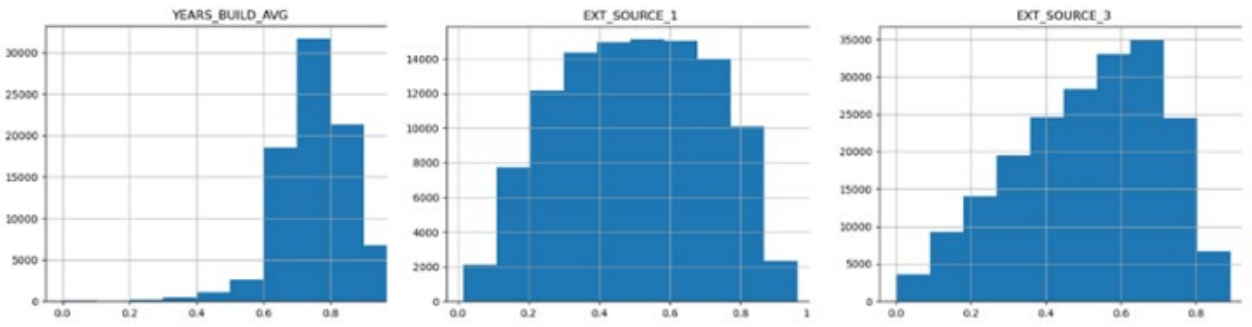

Fig. 2. Distribution of numerical variables - normal distribution

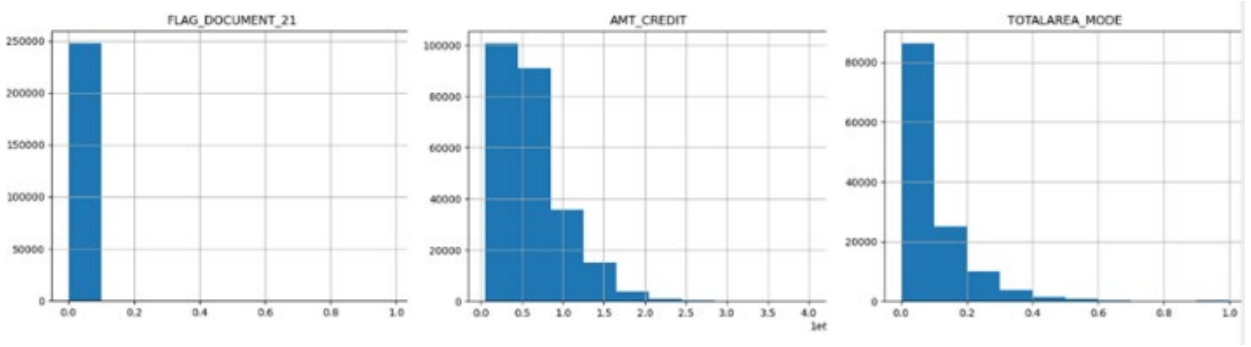

Fig. 3. Distribution of numerical variables - unilateral distribution

The text-based data is discrete, with a total of 16 features, including FLAG_OWN_CAR, EMERGENCYSTATE_MODE, etc. The following Fig. 4 is a partial example of the distribution of text-based data.

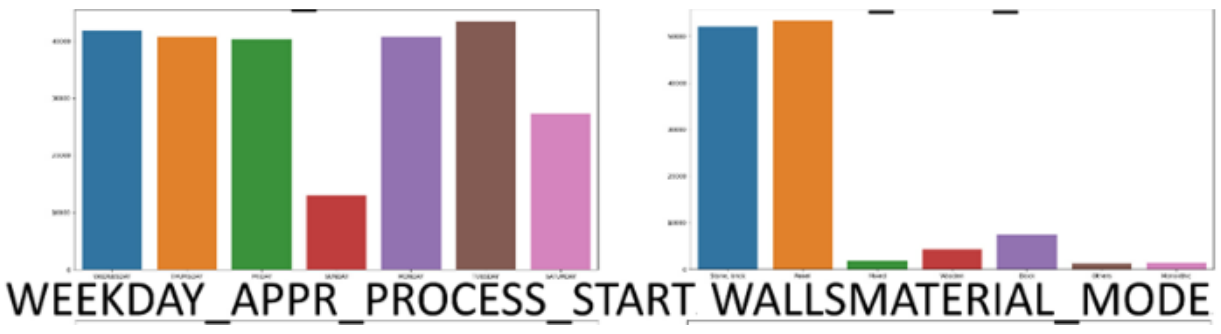

Fig. 4. Distribution of text-based variables

We then classify each feature according to the realistic meaning and calculate the correlation between each other and with TARGET, and classify the features into four categories, which are basic information (as in Fig. 5), standardized information of the customer's residential building (as in Fig. 6), binary variables (as in Fig. 7), and default risk information (as in Fig. 8). As can be seen from the heat map, the basic customer information is less correlated with TARGET (the first column in Fig. 6), and the correlation coefficients between the information are not remarkable, which can respond to each level of information more comprehensively. Fig. 6 shows the standardized information of customer-occupied buildings, which is also less correlated with TARGET, but more correlated with each other, because they respond to the same type of information, only different variables respond to different parts and use different calculation methods. The overall correlation of each binary variable shown in Fig. 7 is small. Inside the information shown in Fig. 8 about the user's possible breach of contract, the correlation of each indicator is small overall, but there is a larger correlation between loan delinquency (OBS_30_CNT_SOCIAL_CIRCLE) and loan default (DEF_30_CNT_SOCIAL _CIRCLE), which is in line with the real situation.

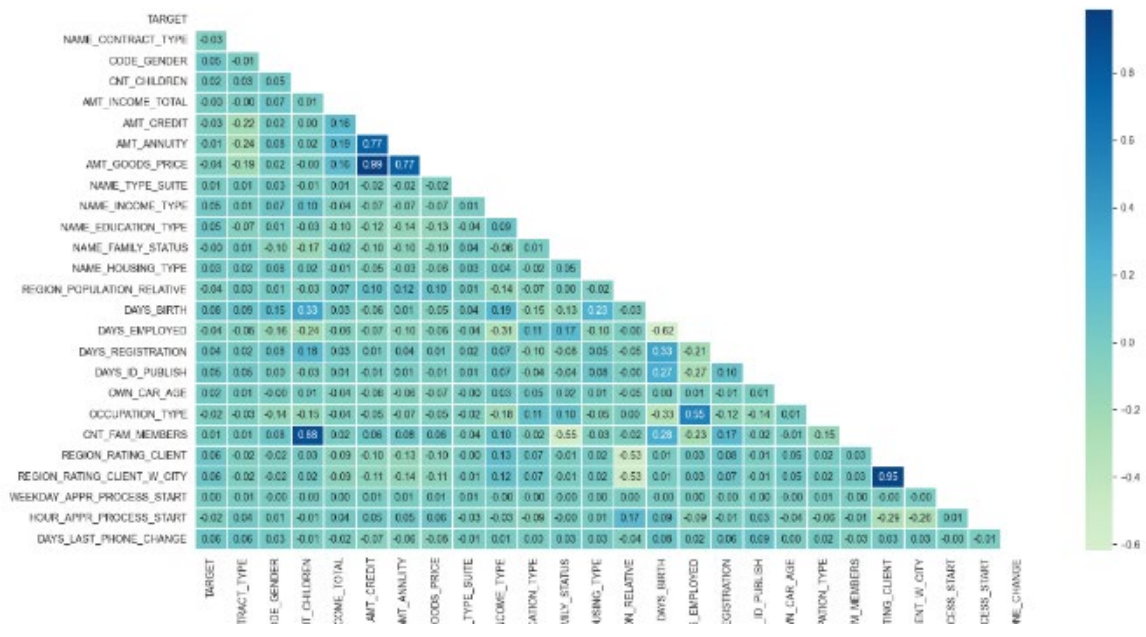

Fig. 5. The correlation of the first part of variable

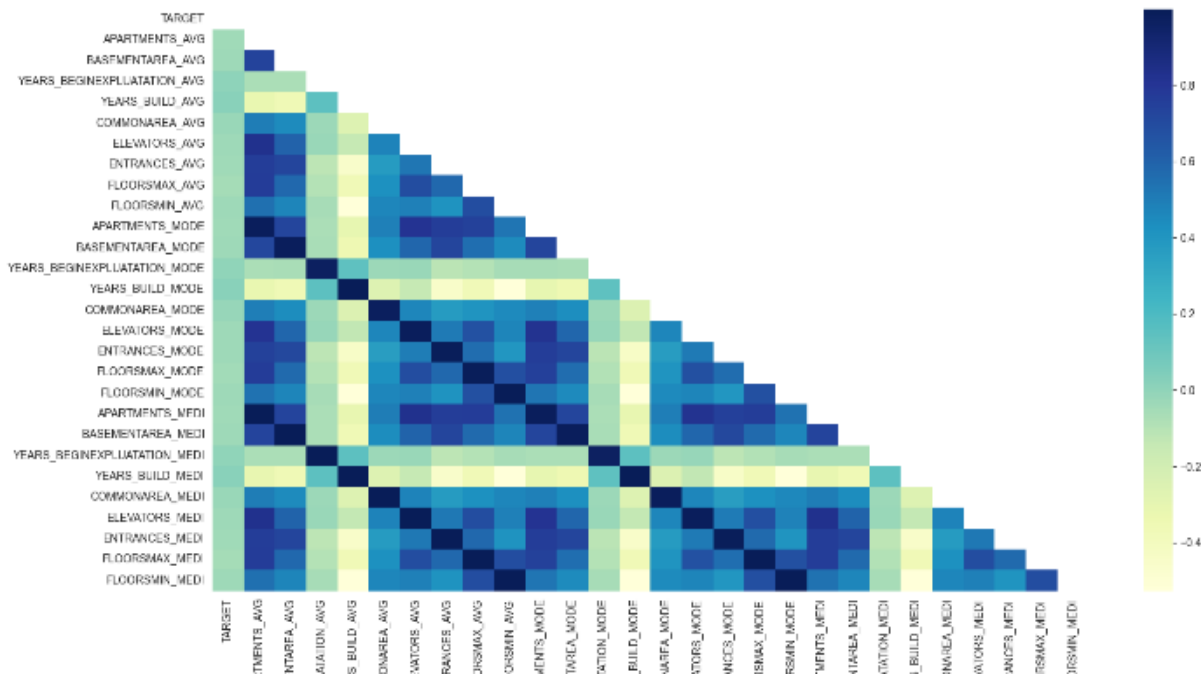

Fig. 6. The correlation of the second part of variable

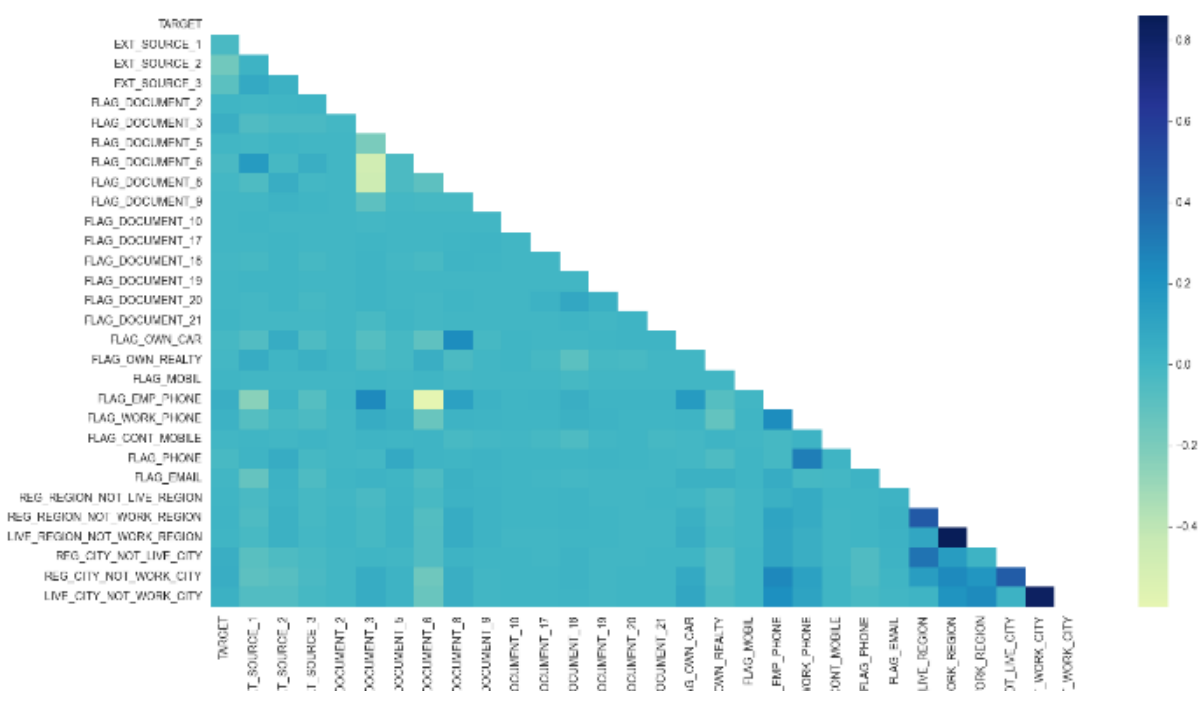

Fig. 7. The correlation of the third part of variable

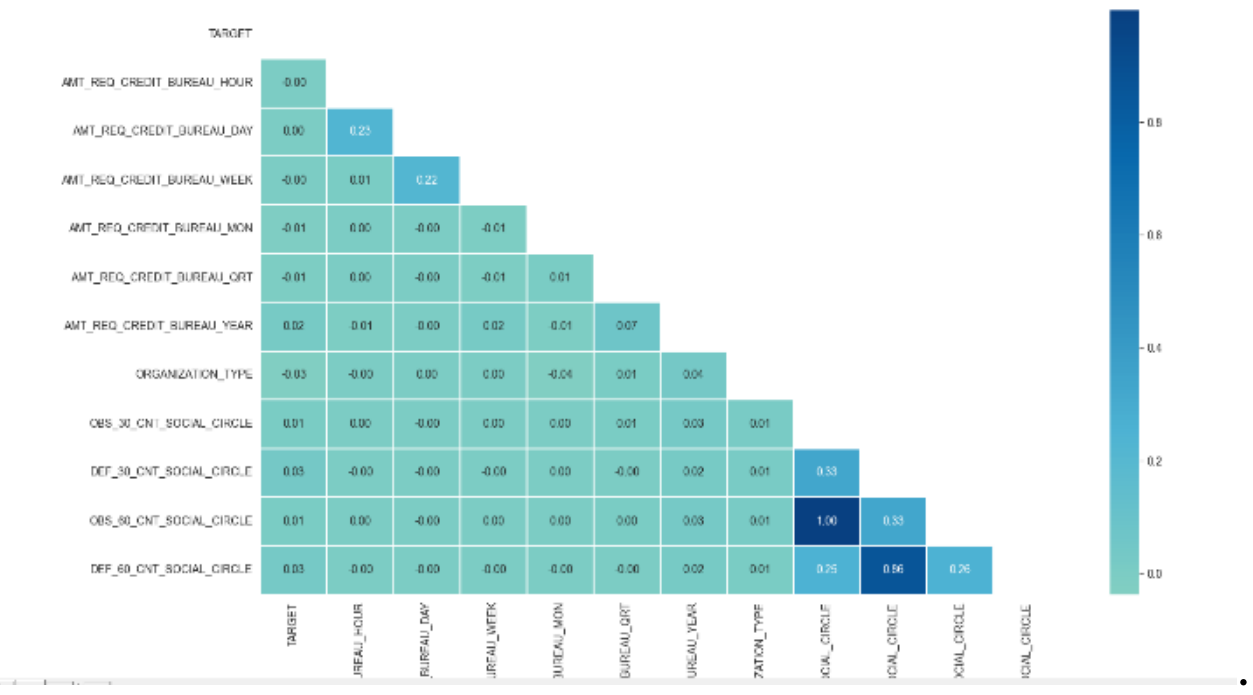

Fig. 8. The correlation of the fourth part of variable

## III. Algorithm Introduction

### *A. Previous Studies*

In recent decades, the field of artificial intelligence has developed rapidly, and there are many artificial intelligence algorithms related to data mining applied in credit risk prediction, such as support vector machine, which is well used in credit risk analysis. Support vector machine (SVM) is a statistical learning method based on the structural risk minimization principle proposed by Vapnik. It implements the structural risk minimization principle by maximizing the margin between two support hyperplanes and transforms the classification problem into a quadratic convex naturalization problem to ensure that the model has a globally optimal solution.

Van Gestel T(2003) constructed a financial early warning model based on the least squares support vector machine, which transformed the quadratic programming problem into a linear equation system to solve so that reduced the difficulty of solving and improved the classification accuracy.

Logistic regression is suitable for the fitting of numerical binary output. It is a classification model. In this paper, it serves as a tool for banks to judge whether there is a possibility of default through customer credit and personal data. In the credit industry, John C Wiginton (1980) was the first one to apply the Logistic model to the credit industry, proving that the model has credible forecasting ability [3].

The idea of the algorithm is as follows: for the input n data to be classified, linearly sum the vectors to obtain the reduced dimension data z. Substitute z into a sigmoid function, and divide it into two categories: 1 and 0 according to the size relationship of function value and 0.5. Next, the locally optimal solution is obtained through iteration, and the parameters are updated.

CHEN T proposed XGBoost(eXtreme Gradient Boosting) in 2014, which is based on the principle of gradient boosted decision tree. XGBoost uses a CART regression tree as a tree-based booster[4]. XGBoost expands the objective function to the second order instead of the first order when performing Taylor expansion. In addition, XGBoost also adds L2 regularization to the cotyledon weights, which improves the performance of the algorithm. The objective function of XGBoost is:

$$Obj = \sum_{i=1}^{n} l(y_i, \hat{y}_i) + \sum_{k=1}^{K} \Omega(f_k) \qquad (1)$$

Where the first term is training loss (squared error, cross-entropy, etc.) and the second term is the regularization loss (L1, L2, etc.).

### *B. LightGBM*

Considering the current state of research, we choose LightGBM as our machine learning algorithm. LightGBM is a scalable machine learning system launched by Microsoft in 2017. It's a distributed gradient lifting framework based on GBDT (Gradient Lifting Decision Tree). It uses GBDT as basic algorithms and adds a series of new functions, such as Histogram Optimization, Vertical Growth Algorithm based on Maximum Depth Leaf-wise, Gradient-based One-Side Sampling (GOSS), and Exclusive Feature Bundling (EFB)[5].

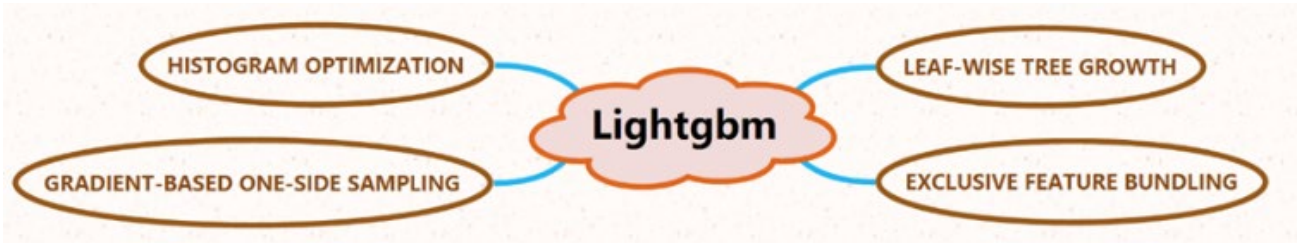

Fig. 9. Lightgbm structure

#### 1) *Histogram Optimization*

The basic imdea of Histogram Optimization is to convert feature value to the bin before training. It divides the continuous features into k discrete features and constructs a histogram with width k to collect information. The histogram of a leaf node can be obtained by subtracting that of its sibling node from the parent node. In practice, we calculate the leaf node which has a smaller histogram, then use subtract to obtain a leaf node with a larger histogram. In this way, Lightgbm can fleetly obtain a histogram containing more data after constructing a histogram with fewer data.

#### 2) *Vertical Growth Algorithm based on Maximum Depth Leaf-wise*

Each time, Lightgbm finds the leaf node with the largest split gain, and cycles until the stop criterion. Using Leaf-wise reduces the computational complexity, and prevents over-fitting in combination with the limitation of the maximum depth. Since the node with the maximum gain needs to be calculated each time, it cannot be split in parallel.

#### 3) *Gradient-based One-Side Sampling*

The smaller the gradient of GBDT algorithm, the better the fitting effect. The GOSS algorithm retains the samples with large gradients and conducts random sampling on the samples with small gradients, so as to reduce the number of samples with small gradients and greatly reduce the computational burden. The main functions of this algorithm include that: ① It can better fit under-trained samples; ② Multiply the weight to reduce the influence of sampling on the distribution of the original data.

*4) Exclusive Feature Bundling*

EFB is a nearly lossless method to reduce the size of the dataset by cutting down the number of features. For a dataset containing a large number of features, many of the features may be exclusive, which implies they rarely take nonzero values simultaneously, such as one-hot encoded features. In this algorithm, these features are bundled together to form a new feature to reduce the number of features and improve the training speed

## IV. Experiment

### A. Feature Engineering

*1) Missing Data Imputation*

We counted the missing data of each label and found that there is missing data under 64 labels, among which there were 41 tags with 50% missing data, and 14 tags with 20% missing data. Thus it is necessary to fill up that missing data before training the model in the next stage.

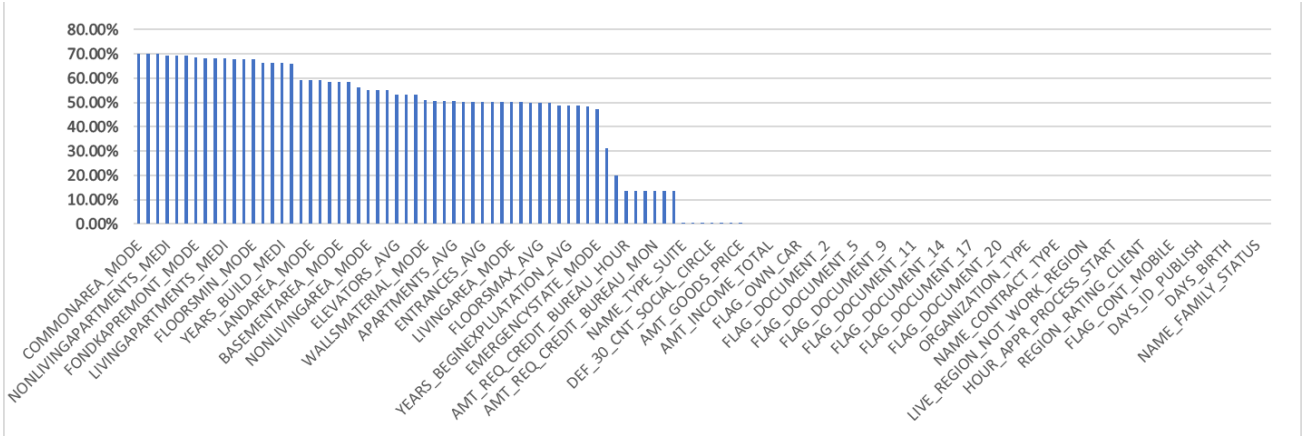

Fig. 10. Proportion of missing data

Since the LightGBM algorithm we used is less sensitive to missing values, in order to cover as many feature variables as possible in the model training, we kept all the feature values, divided the feature variables into two categories, textual and numerical, and processed the missing values using different filling methods and got different dataset. We used them to train the model and compared the AUC that different models generated(Table Ⅱ). We perform separate column operations on the data under each label. While we fill the missing value of textual type with mode or just keep it, we fill the missing value of numerical type with median, round mean, mode, and zero. Many numerical variables have integer values, so we rounded the mean value. And a significant number of variables have the value 0, so zero-imputation is an appropriate strategy.

In conclusion, with other parameters unchanged, the AUC generated by the model that used the plural to fill the textual type data and used the mean rounding to fill the numerical type data is the highest, reaching 0.7631, and method 6 is chosen as our missing value treatment method.

Table II. Missing value imputation method

| *Method Number* | *Textual Type* | *Numerical Type* | *AUC* |
|---|---|---|---|
| 1 | do nothing | median | 0.7621 |
| 2 | do nothing | round mean value | 0.7625 |
| 3 | do nothing | mode | 0.7624 |
| 4 | do nothing | zero | 0.7618 |
| 5 | mode | median | 0.7616 |
| 6 | mode | round mean value | 0.7631 |
| 7 | mode | mode | 0.7621 |
| 8 | mode | zero | 0.762 |

*2) Encoding using the LabelEncoder method*

In order to perform other operations, we encode the textual data. The encoding means that the nominal variable values are assigned different numbers according to different categories. We choose to use the LabelEncoder method, which encodes the variable values of textual descriptions using consecutive natural numbers (0,1,2,3 ......n-1,n is the variable type number). Since LightGBM comes with category processing, we do not choose to continue using the one-hot encoder method. Unlike simple one-hot coding, LightGBM can find the optimal segmentation of categorical features and provide more accurate optimal segmentation compared to one-hot coding results. LightGBM provides good accuracy when using local classification features.

*3) Standardization and normalization*

In conventional machine learning and data mining, the normalization of data is an important method for data pre-processing to improve the accuracy of machine learning. Normalization generally maps the data to a specified range, which is used to remove the magnitudes as well as the units of magnitudes of different dimensional data. The common mapping ranges are [0, 1] and [-1, 1], and the most common normalization method is Min-Max Normalization. Min-Max normalization is a linear transformation of the original data so that the resultant values are mapped between [0,1]. The conversion function is as follows:

$$x_{new} = \frac{x - x_{min}}{x_{max} - x_{min}} \quad (2)$$

Where $x_{max}$ is the maximum value of the sample data and $x_{min}$ is the minimum value of the sample data. This normalization method is more suitable in the case of relatively concentrated values. However, if $x_{max}$ and $x_{min}$ are unstable, it is easy to make the normalization result unstable, making the subsequent use of the effect also unstable.

Standardization is to process the data according to the columns of the feature matrix. There are various methods of data normalization, such as linear methods, dash-line methods, and curvilinear methods. Different standardization methods can have different effects on the evaluation results of the model. Among them, the most commonly used is Z-Score Normalization. Z-Score Normalization gives the mean and standard deviation of the original data to standardize the data. The processed data conform to the standard normal distribution, i.e., the mean is 0 and the standard deviation is 1. The transformation function is:

$$x_{new} = \frac{x - \mu}{\sigma} \quad (3)$$

Where μ is the mean of the sample data and σ is the standard deviation of the sample data. In addition, the normalized data keeps the useful information in the outliers, making the algorithm less sensitive to the outliers, which normalization cannot be achieved. We normalize and standardize the data and get two datasets that were used to train the model separately. Then compare two models with the model trained on raw data. By comparing the accuracy, precision, and AUC of the test set, we can select the most appropriate strategy for data preprocessing.

Table III. COMPARISON OF THE RESULTS GENERATED BY DIFFERENT MODELS

| | *Accuracy* | *Precison* | *AUC* |
|---|---|---|---|
| Raw data | 0.7259 | 0.175 | 0.7624 |
| Normalized data | 0.7211 | 0.1739 | 0.7617 |
| Sandardized data | 0.7255 | 0.174 | 0.7622 |

As shown in Table III, comparing the performance of three LightGBM models, we can see the difference between the results to which the three processing methods lead is relatively small, for that the overall AUC is around 0.762, the accuracy and precision are similar, and the precision of all three methods is only 0.17. In conclusion, the AUC of the results without normalization and normalization is slightly higher, and the overall training time of the model can be reduced. Thus, we decided not to normalize and normalize the data during the training of LightGBM model.

*4) Imbalanced dataset*

The dataset we use has a total of 307,511 records with two categories. 282,686 records are "0", which means there are 282,686 records of customers who do not default, 24,825 records are "1", which means customers who default. The ratio of positive and negative samples is about 1:11.4, the dataset is imbalanced.

The methods to deal with imbalanced dataset mainly includes Over-sampling and under-sampling. Over-sampling is a method that randomly samples and replicates the minority sample set without losing the original information to keep the overall sample proportion balanced. Under-sampling is to randomly delete the majority of class samples to achieve the purpose of balancing the proportion of the two types of data sets. Because under-sampling will delete most samples, and the data set we use has a great difference in the proportion of positive and negative samples, using under-sampling will lose a lot of data, so we chose the improved algorithm ADASYN of the Over-sampling method.

ADASYN: adaptive synthetic sampling, it gives different weights to different minority samples to generate different numbers of samples. The specific process is as follows[6]:

*a) Calculate the number of synthetic data examples that need to be generated for the minority class:*

$$G = (m_l - m_s) \times \beta \tag{4}$$

where $m_l$ is the number of samples of most classes, $m_s$ is the number of minority samples, and $\beta \in [0,1]$ is a random number. If β is equal to 1, the positive-to-negative ratio after sampling is 1:1.

*b) Calculate the proportion of most classes in K nearest neighbors:*

$$r_i = \Delta_i / K \tag{5}$$

where $\Delta_i$ is the number of samples of most classes in K nearest neighbors， $i = 1,2,3 \ldots m_s$

*c) Normalize $r_i$ according to*

$$\hat{r_i} / \sum_{i=1}^{m_s} r_i$$

*d) Calculate the number of new samples to be generated for each minority sample according to the sample weight:*

$$g = \hat{r_i} \times G \tag{6}$$

*e) Calculate the number of samples to be generated for every few samples according to $g$, and generate samples*

$$s_i = x_i + (x_{zi} - x_i) \times \lambda \tag{7}$$

where $s_i$ is a generated sample, $x_i$ is the i-th sample in a minority, $(x_{zi} - x_i)$ is the difference vector in n-dimensional spaces, and λ is a random number: $\lambda \in [0,1]$

We will use ADASYN to deal with imbalanced data in the comparison algorithm Logisics and SVM. XGboost and LightGBM have their imbalanced sample processing methods, so we use the built-in algorithm to deal with imbalanced data. LightGBM improves the calculation weight of a small number of samples to achieve the same influence of a small number of samples and a large number of samples. Set LightGBM parameter *is-unbalance=True* to balance positive and negative samples during model training.

*5) Create new properties*

First, we use the initial parameters to set the LightGBM algorithm, use the remaining 120 attributes for model training, and set the number of training n_ estimators=5000. After that, calculate the information gain of each attribute. Information gain is used to measure the gain of an attribute after splitting it. The calculation formula is as follows

$$Gain = \frac{1}{2}\left[\frac{G_L^2}{H_L + \lambda} + \frac{G_R^2}{H_R + \lambda} - \frac{(G_L + G_R)^2}{H_L + H_R + \lambda}\right] - \gamma \tag{8}$$

Where λ is the second-order regularization coefficient, γ is the first-order regularization coefficient, $G_L^2$ represents the sum of the first derivative of the left subtree.

The AUC of the model is 0.762. Sort top 20 attributes according to the information gain, and draw the importance histogram of each attribute, as shown in Fig. 11.

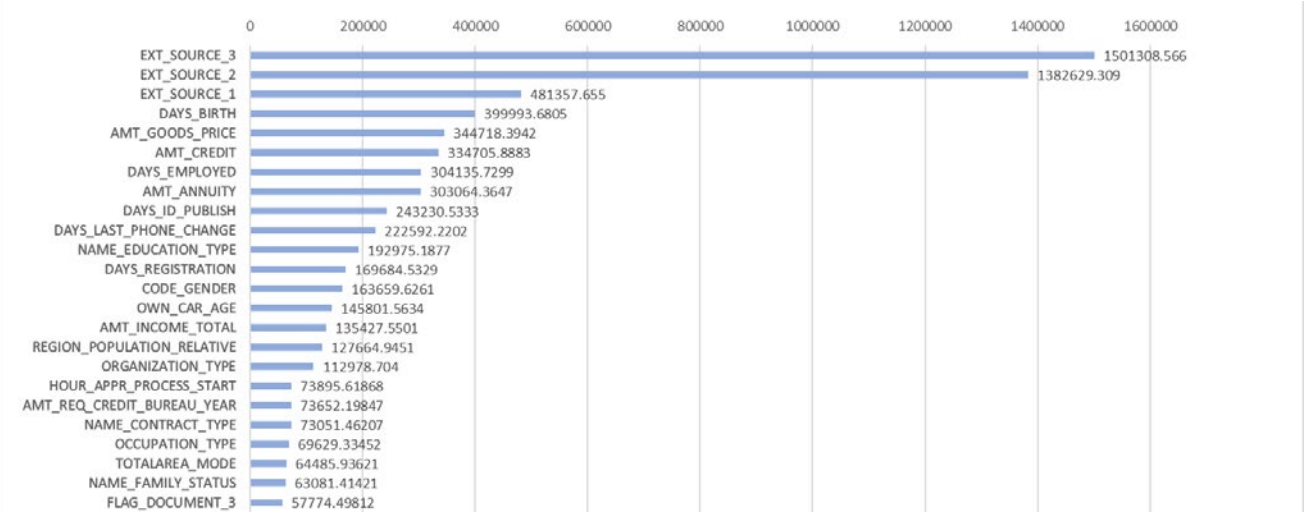

Fig. 11. Attribute Importance Ranking

It can be seen from Fig. 11 that in LightGBM model training, EXT_ SOURCE_ 3、EXT_ SOURCE_ 2 and EXT_ SOURCE_ 1 which represent the standardized scores of external data sources have the highest scores . It can be seen that other external data has a great impact on the model effect.

To improve the prediction effect of the model, we use division and addition to construct five new attributes for training according to the importance of attributes and the natural meaning of attributes, as shown in Table Ⅳ. The scores of these newly constructed attributes in the model are shown in Fig. 12. It can be seen that the scores of the five new attributes are all in the top 20, and the overall score is high, and credit_ annuity_ ratio score exceeds EXT_ SOURCE_ 1 becomes the third most important attribute among 124 attributes. Using 124 attributes for training, the AUC of the model has reached 0.7715, and the five new attributes have significantly improved the prediction effect of the model.

Table IV. FIVE NEW CONSTRUCTION PROPERTIES AND THEIR MEANINGS

| ***Number*** | ***Attribute Name*** | ***Construction method*** |
|---|---|---|
| 1 | cerdit_annuity_ratio | AMT_CREDIT/ AMT_ANNUITY |
| 2 | prices_income_ratio | AMT_GOODS_PRICE/AMT_INCOME_TOTAL |
| 3 | employed_age_ratio | DAYS_EMPLOYED/ DAYS_BIRTH |
| 4 | credit_goods_ratio | AMT_CREDIT / AMT_GOODS_PRICE |
| 5 | ext_source_sum | EXT_SOURCE_3+EXT_SOURCE_2+EXT_SOURCE_1 |

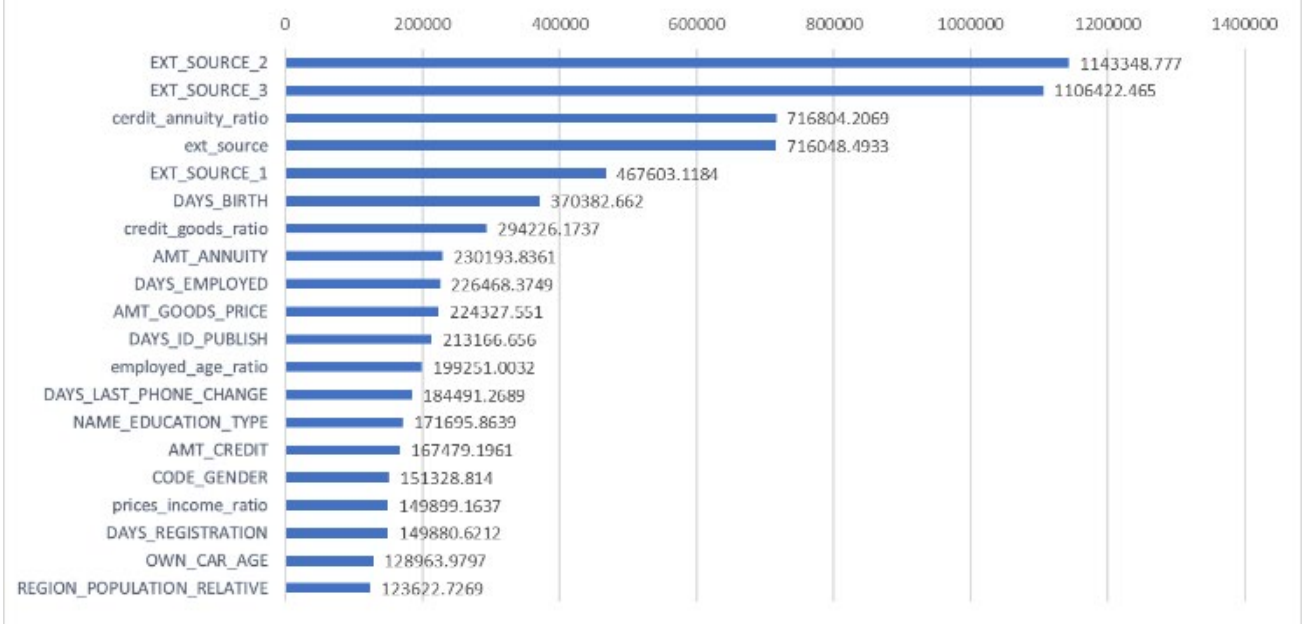

Fig. 12. The importance of scoring the top 20 attributes

To improve the prediction effect of the model, we use k-means clustering algorithm to cluster the top 5 attributes into 10 categories and use these categories as a new attribute classification for model training.

Finally, the data set we used included the initial 119 attributes and six attributes constructed manually, totaling 125 attributes. The new data set was used for model training.

## *B. Model Training*

### *1) Data set division*

The total dataset has 307,511 records, which we divided by random sampling into a training set with 246,008 records and a test set with 61,503 records. The ratio of training set to test set is 4:1.

### *2) Parameter Optimization*

The main LightGBM parameters are *boosting_type*, *colsample_bytree*, *subsample*, *subsample_for_bin*, *min_split_gain*, *learning_rate*, *num_leaves*, and *max_depth*, *reg_alpha*, *reg_lambda*, *is_unbalance*.

For most of the LightGBM parameters, we used the GridSearchCV method with a fixed *learning_rate*. The principle of this method is to loop through the parameters to be tuned, compare the training results of the model with different parameter combinations, and select the best parameter combination. This method uses cross-validation in the process of comparing the training results. *num_leaves* is constrained by *max_depth*, which is the main parameter used to control the complexity of the tree model. We combined the two and tuned them. The optimal result was *max_depth=6* and *num_leaves=50*. We used the same method to obtain the optimal values of 0.44 and 0.48 for *reg_alpha* and *reg_lambda*. Both prevent the model from over-fitting.

The *boosting_type* parameter determines the type of algorithm used for boosting, and we chose the gradient-based one-sided sampling algorithm. *colsample_bytree* can be used to speed up training and deal with overfitting by controlling how many features are randomly selected at each iteration. The *subsample* is similar to *colsample_bytree*. *subsample_for_bin* determines the amount of data used to build the histogram. The sparser the data, the larger the value should be set to get better training results and increase the data loading time. *min_split_gain* specifies the minimum gain to perform the split, which affects the depth of the tree. A smaller value will ensure the accuracy of the model prediction but may cause over-fitting, so we found the optimal value of 0.03023 through traversal.

The parameter *is_unbalance* indicates whether the sample is balanced. If *is_unbalance=True*, it means that the sample is unbalanced, and LightGBM will adjust the weight of minority samples to achieve sample balance.

*learning_rate* means the rate of shrinkage of the model and affects the normalized weight of the dropped tree. The lower the *learning_rate*, the slower the model trains. Because the dataset is large and has many features, we set it to a high value so that we can adjust the rest of the parameters, and then reduce it to a suitable value. We chose *learning_rate = 0.0545*.

Table V. LIGHTGBM PARAMETER SETTINGS

| ***Parameter Name*** | ***Value*** | ***Parameter Name*** | ***Value*** |
|---|---|---|---|
| colsample_bytree | 0.4888 | num_leaves | 50 |
| subsample | 1 | max_depth | 6 |
| subsample_for_bin | 240000 | reg_alpha | 0.44 |
| min_split_gain | 0.03023 | reg_lambda | 0.48 |
| learning_rate | 0.0545 | boosting_type | goss |

The LightGBM with adjusted parameters was used for training, and the number of training arguments was set to 5000 to obtain the test AUC change graph, as shown in Fig. 13. It can be seen from the figure that the AUC increases rapidly in the first 100 rounds, and then increases slowly, reaching the peak at 3521 rounds with AUC= 0.771908. After that, it is over-fitting and AUC decreases slightly. The model with the peak AUC was selected as the classifier for model evaluation.

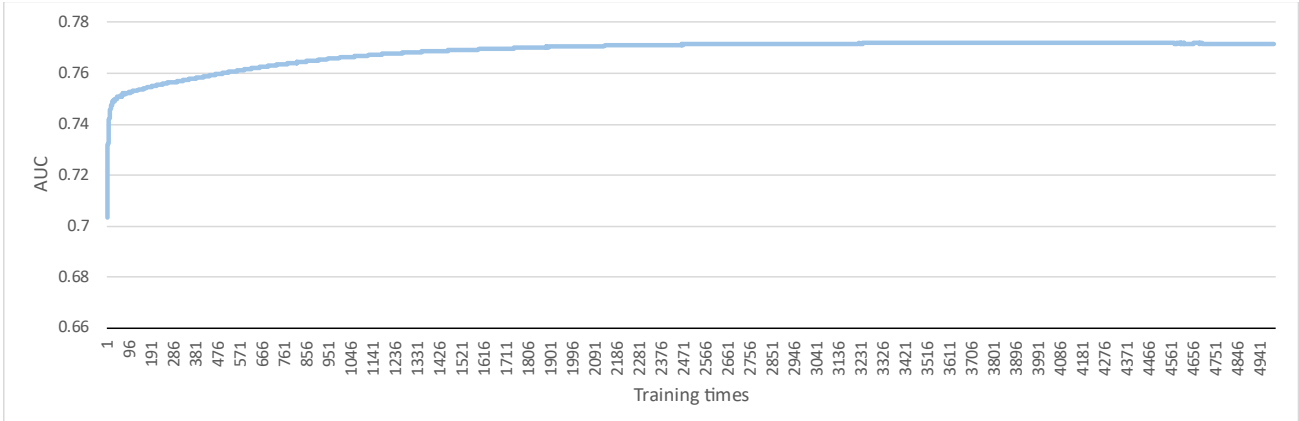

Fig. 13. AUC of LightGBM models varies with the training times

## *C. Model Evaluation*

### *1) Evaluation Parameters*

Table VI. CONFUSION MATRIX

| **Confusion matrix** | | ***Real class*** | |
|---|---|---|---|
| | | T | F |
| ***Predicted class*** | T | True Positive | False Positive |
| | F | False Negative | False Negative |

Accuracy: The proportion of samples with the correct classification.

$$Accuracy = \frac{TP+TN}{TP+TN+FP+FN} \quad (9)$$

Precision: It indicates how many of the predicted positive samples are true positive samples.

$$Precision = \frac{TP}{TP+FP} \quad (10)$$

Recall: It indicates how many positive examples in the sample are classified correctly.

$$Recall = \frac{TP}{TP+FN} \quad (11)$$

F1-score: F1 score can be regarded as a harmonic average of model accuracy and recall, with the maximum value of 1 and the minimum value of 0.

$$F_1 = 2 \times \frac{Precision \times Recall}{Precision + Recall} \quad (12)$$

ROC: Receiver Operating Characteristic Curve, is a comprehensive indicator of continuous variables of response sensitivity and specificity. Its abscissa and ordinate are respectively False Positive and True Positive. When the number of true and false positive cases in the prediction result is equal, the ROC curve is $y = x$. When drawing curves, suppose that there are $m^+$ real cases and $m^-$ false positive cases in the sample prediction results. Starting from the (0,0) point, let the coordinates of the previously marked point be (x, y). When a true positive case occurs, mark the point at $\left(\frac{x+1}{m^+}, y\right)$. When a false positive case occurs, mark the point at $\left(x, \frac{y+1}{m^-}\right)$. Finally, connect the marked points with a smooth curve to get the ROC curve.

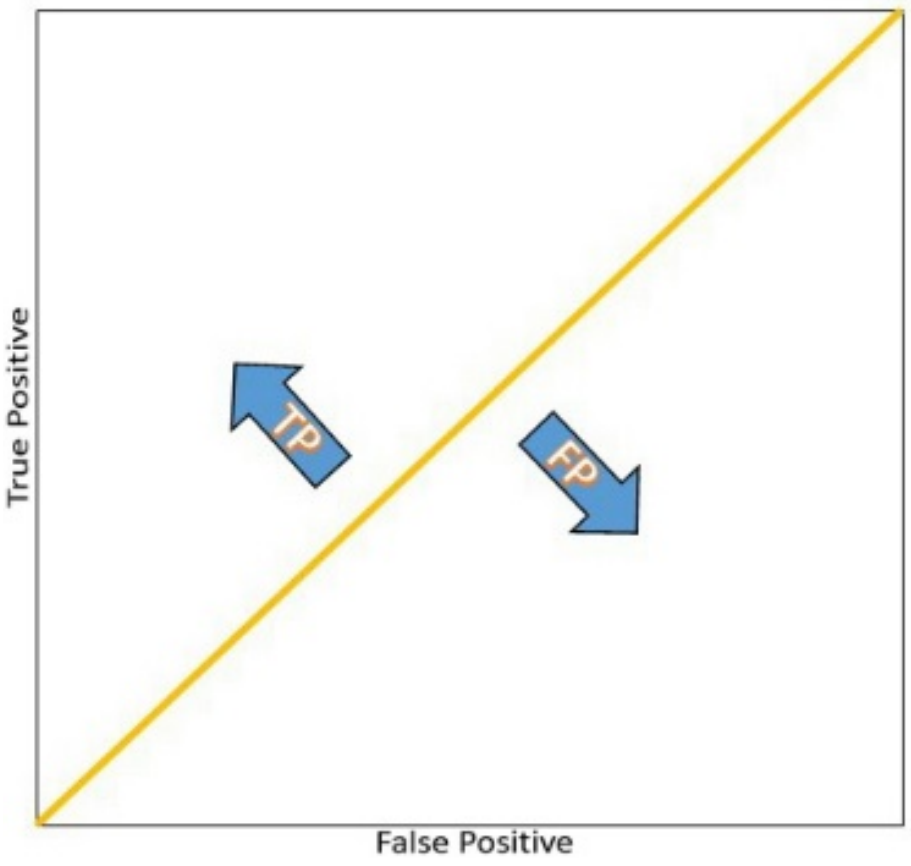

Fig. 14. ROC curve plotting process

AUC: It is the area under the ROC curve. Obviously, the larger the AUC is, the better the prediction effect of the model is.

*2) Model Evaluation*

LightGBM sets 5000 rounds of training and selects the optimal AUC as the result of classifier. XGboost sets learning parameters similar to LightGBM. Logistic optimization parameters select *solver= 'lbfgs'*. Considering the limitation of computing power and the characteristics of SVM algorithm, we selected 10% of the original data set for SVM training and testing. The results of each model are shown in Table Ⅶ.

Table VII. TEST RESULTS OF EACH CLASSIFIER. THE TIME IS TAKEN AS THE AVERAGE OF FIVE CONSECUTIVE TRAINING.

| | *Accuracy* | *Precison* | *Recall* | *F1 score* | *AUC* | *Time(s)* |
|---|---|---|---|---|---|---|
| Lightgbm | 0.734 | 0.181 | 0.657 | 0.284 | 0.772 | 213 |
| XGBoost | 0.849 | 0.235 | 0.39 | 0.293 | 0.735 | 2421 |
| Logistic | 0.645 | 0.544 | 0.641 | 0.589 | 0.692 | 251 |
| SVM | 0.658 | 0.543 | 0.637 | 0.586 | 0.688 | [a] |

a.Since the data set used in SVM training and testing is different from other classifiers, the time of SVM is not calculated here.

As can be seen from Table VII, XGboost performs best in accuracy, reaching 0.849, precision but recall are low. LightGBM has the highest recall and AUC that reaches 0.772, ranking second in accuracy. Logist and SVM have similar performance in various indicators, and the overall performance is weaker than other algorithms. In terms of efficiency, LightGBM performs best among the four classifiers with 213s running time. Logistic benefits from the fast running speed of linear mode. SVM is extremely inefficient because it is not suitable for large samples. To sum up, LightGBM after parameter optimization has the best prediction effect on whether customers may default, and its AUC is relatively high, which can provide some reference for banks to provide customers with credit loans.

## V. CONCLUSION

This paper is mainly based on the customer information data set of a foreign commercial bank in Kaggle, and we use machine learning algorithm to build a classifier to classify customers into two categories. The classification is used to determine the possibility of user default, so as to help the bank better judge whether to give customers credit loans. In this paper, LightGBM algorithm is used in machine learning, helping to process the data sets, such as missing value processing, coding and constructing attributes, etc., which improved the effect of the algorithm. Finally, the accuracy of LightGBM classifier exceeds 0.7, while AUC reaches 0.77. Compared with classifiers constructed by other classifiers, the LightGBM classifier has obvious advantages in AUC and running efficiency, which can better classify the default behavior of commercial bank customers, and this classifier has a certain value for risk management.